\documentclass[10pt,twocolumn,letterpaper]{article}

\usepackage[pagenumbers]{wacv} %

\usepackage{graphicx}
\usepackage{amsmath}
\usepackage{amssymb}
\usepackage{booktabs}
\usepackage{algorithm}
\usepackage{dsfont}
\usepackage{algorithm}
\usepackage{algpseudocode}
\usepackage{algorithmicx}
\usepackage{mathtools}
\usepackage{multirow}
\usepackage{graphics}
\usepackage{wrapfig}
\usepackage{bbm}
\usepackage{xspace}
\usepackage{xcolor}
\usepackage{tikz}

\newcommand{\methodName}{RapidMV\xspace}

\usepackage{xr}
\makeatletter
\newcommand*{\addFileDependency}[1]{%
  \typeout{(#1)}
  \@addtofilelist{#1}
  \IfFileExists{#1}{}{\typeout{No file #1.}}
}
\makeatother

\definecolor{wacvblue}{rgb}{0.21,0.49,0.74}
\usepackage[pagebackref,breaklinks,colorlinks,allcolors=wacvblue]{hyperref}

\def\wacvPaperID{862} %
\def\confName{WACV}
\def\confYear{2026}

\title{RapidMV: Leveraging Spatio-Angular Latent Space for Efficient and Consistent Text-to-Multi-View Synthesis
}

\author{Seungwook Kim$^{1,2}$ \hspace{0.6cm} Yichun Shi$^2$ \hspace{0.6cm} Kejie Li$^3$ \hspace{0.6cm} Minsu Cho$^1$ \hspace{0.6cm} Peng Wang$^2$ \vspace{0.3cm}\\
$^1$POSTECH, South Korea \hspace{1.5cm} $^2$ByteDance Seed, USA \hspace{1.5cm} $^3$Meta, USA 
}

\begin{document}
\twocolumn[{
\renewcommand\twocolumn[1][]{#1}%
\maketitle
\vspace{-6.0mm}
\includegraphics[width=\textwidth]{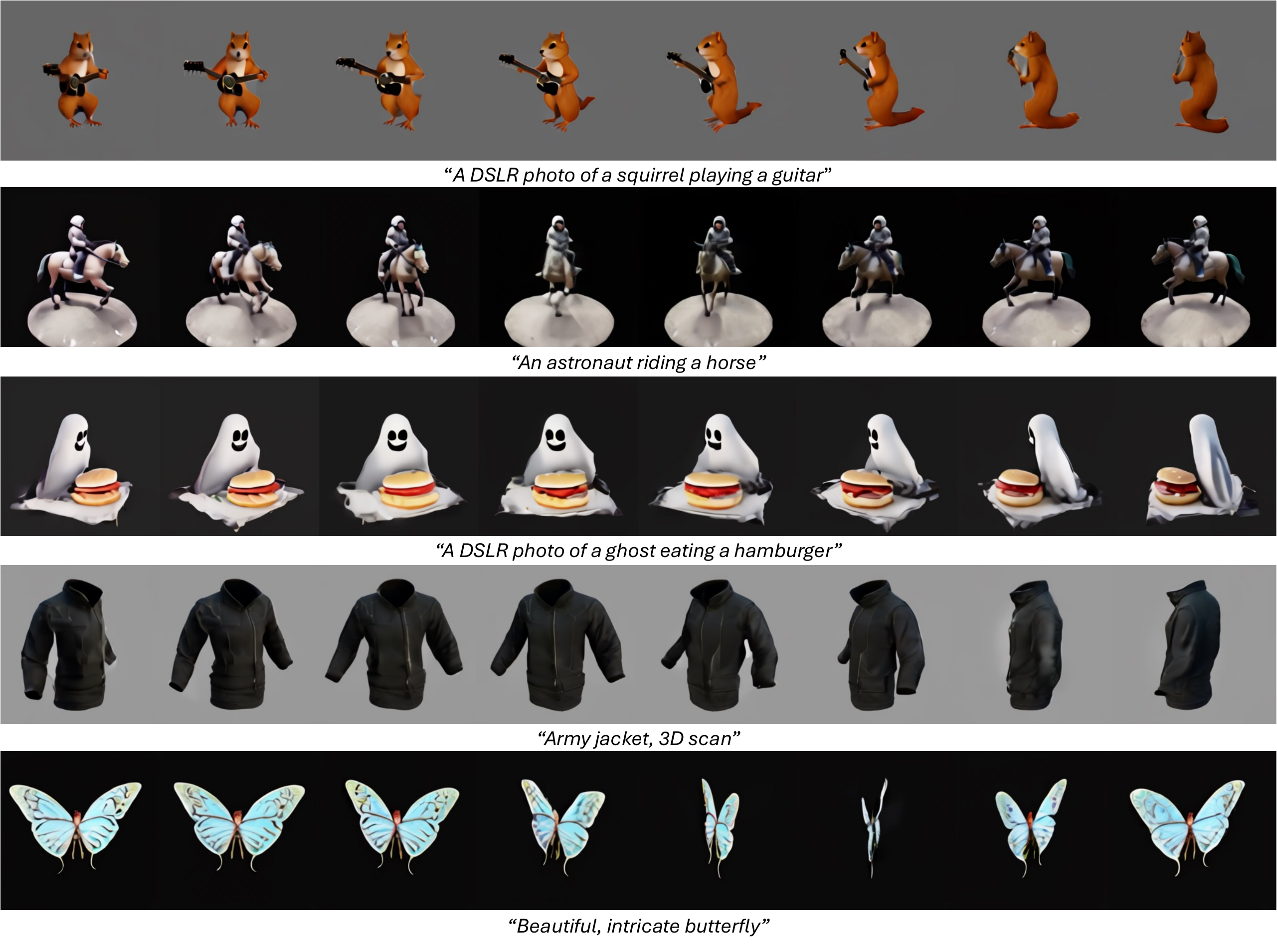}
\vspace{-6.5mm}
\captionof{figure}{\textbf{Multi-view generation results of \methodName.}
We visualize 8 frames from 32 generated views of \methodName.
\methodName can generate 32 multi-view images for a given text prompt in just around 5 seconds.
} 
\vspace{+3.0mm}
\label{fig:teaser}
}]

\begin{abstract}
Generating consistent multi-view images given a text prompt is an essential bridge to generating synthetic 3D assets.
In this work, we introduce \methodName, a novel text-to-multi-view generative model that can produce 32 multi-view synthetic images in just around 5 seconds.
In essence, we introduce a novel spatio-angular latent space, where we encode not only the spatial appearance of a single frame, but also the angular viewpoint deviations across multiple frames into a single latent for improved efficiency and multi-view consistency.
We achieve effective training of \methodName by strategically decomposing our training process into multiple steps.
We demonstrate that \methodName outperforms existing methods in terms of consistency and latency, with competitive quality and text-image alignment.
\end{abstract}
    
\section{Introduction}
\label{sec:intro}

Recent advances in text-to-image generation~\cite{nichol2022glide, rombach2022high} have been driven by the generative capabilities of diffusion models~\cite{nichol2021improved, ho2020denoising} and the availability of large-scale text-image paired datasets~\cite{schuhmann2022laion, gadre2024datacomp}.
Text-to-multi-view generation extends this paradigm by aiming to generate \textit{multiple} views of an object described by a textual prompt, with each view depicting the object from a different viewpoint.
This multi-view generative capability serves as a crucial bridge toward 3D generation, where the generated multi-view images provide the necessary cues to ultimately generate or reconstruct a 3D asset~\cite{poole2022dreamfusion, shi2023mvdream, kim2024correspondentdream, lin2023magic3d, wang2023prolificdreamer}.

However, existing methods show limitations in (1) \textbf{the density of multi-view images}, \ie, the number of multi-view images that can be generated for a text prompt , and (2) \textbf{the multi-view consistency} between the generated views -- which are both vital aspects of bridging multi-view generation to 3D generation. 
Lack of dense multi-view images introduces reliance on computation-heavy algorithms~\cite{poole2022dreamfusion} or further generative networks~\cite{tang2025lgm, li2023instant3d} for 3D generation, while lack of multi-view consistency can result in dissatisfactory results in the generated or reconstructed 3D output~\cite{shi2023mvdream, tang2025lgm}.
This task is further complicated by the limited scalability and diversity of existing text-3D paired datasets, which hinders the development of models capable of handling a wide range of objects and viewpoints.

In this work, we present \methodName, a strong diffusion-based text-to-multi-view generation method that produces high-quality, consistent, and dense multi-view images.
To facilitate this, we propose the new \textbf{spatio-angular latent space}, which encodes not only the spatial appearance of a single frame, but also the angular viewpoint deviations across multiple frames into a single latent; exhibiting improved multi-view consistency across frames. 
To enable spatio-angular latents to attend to each other for improved consistency in angular deviation and appearance generation, we introduce the \textbf{Global Spatio-Angular Attention}.
To generate spatio-angular latents from desired viewpoints, we incorporate the \textbf{Latent-wise Anchor-pose Modulation} to modulate the diffusion model with respect to the desired viewpoint \ie, camera pose, we want to base our generation upon.
Ultimately, the use of spatio-angular latents dramatically boosts the efficiency of generation, enabling \methodName to generate 32 multi-view images of an object in just around 5 seconds, a significant boost in comparison to existing methods, which take at least $\sim40$ seconds.

We quantitatively evaluate \methodName against existing text-to-multi-view methods, demonstrating state-of-the-art consistency and latency, with competitive quality and text-image alignment.
A comprehensive user study shows that \methodName generates more appealing multi-view images of an object compared to existing methods.

Our contributions are fourfold:

\begin{itemize} 
\item We introduce \methodName, a novel text-to-multiview diffusion model capable of efficiently generating high-quality, consistent, and dense multi-view images from text.
\item We introduce the new \textit{spatio-angular} latent space to encode both the spatial appearance and angular viewpoint deviations, which significantly enhances the latency and consistency of the multi-view generation process.
\item We propose two key modules, the Global Spatio-Angular Attention and the Latent-wise Anchor-Pose Modulation, to effectively leverage and manipulate spatio-angular latents to generate high-quality multi-view outputs.
\item Through qualitative evaluation and a comprehensive user study, we validate that \methodName outperforms existing methods in terms of consistency and latency, while exhibiting competitive quality and text-image alignment.
\end{itemize}

\section{Related Work}
\label{sec:relatedwork}

\smallbreak
\noindent
\textbf{Text-to-multi-view generation.}
Following the success of diffusion models in image generation~\cite{song2020denoising, ho2020denoising}, the latent diffusion model~\cite{rombach2022high} propelled the advancement in text-to-image (\ie text-to-single-view) generation, leveraging the latent space for efficient and effective generation~\cite{nichol2022glide, ramesh2022hierarchical, esser2024sd3}.
Building on the success of text-to-single-view diffusion models, MVDream~\cite{shi2023mvdream} first proposed a text-to-multi-view diffusion model, capable of simultaneously generating 4 orthogonal views of an object.
The ability to generate multi-view images led to dramatic improvements in the field of text-to-3D generation; using MVDream's text-to-multi-view diffusion model for SDS~\cite{poole2022dreamfusion, lin2023magic3d, wang2023prolificdreamer}instead of text-to-single-view diffusion models significantly reduces 3D inconsistencies \ie the Janus face or content drift problems, benefiting from the multi-view priors.
Building on MVDream, VideoMV~\cite{zuo2024videomv} finetunes a tex-to-video diffusion model to generate 24 views of an object, achieving enhanced consistency through 3D-aware denoising sampling.
Concurrently, Bootstrap3D~\cite{sun2024bootstrap3d} focuses on improving 4-view generation quality by creating a large-scale synthetic multi-view dataset with dense captions.

However, existing text-to-multi-view diffusion models are limited in terms of (1) density \ie the number of multi-view images they can generate and (2) the multi-view consistency of the generated images. 
Due to these limitations, existing methods have to incorporate the computation-heavy SDS algorithm~\cite{poole2022dreamfusion} or additional generative models~\cite{hong2023lrm, tang2025lgm} to yield the final 3D asset, instead of relying on direct reconstruction using the multi-view images~\cite{mildenhall2021nerf, wang2021neus}. 
In this work, we introduce \methodName, which generates \textbf{32} multi-view images from text, surpassing existing methods in all areas of density, consistency and efficiency.

\smallbreak
\noindent
\textbf{Latent space for diffusion models.}
Stable Diffusion~\cite{rombach2022high} first proposed to operate in the spatial latent space, benefitting from the $8\times8$ spatial compression to enable high-resolution image generation while supporting conditional image synthesis.
This approach required the use of 2D image VAEs~\cite{kingma2013auto, rombach2022high, podell2023sdxl} to compress images into latent representations that could be decoded back to the original image with minimal loss of fidelity.
Building on the success of using the spatial latent space in image generation, initial video diffusion models~\cite{singer2022make, blattmann2023align} also operate in the spatial latent space to encode each frame of the video into a spatial latent.
However, the use of spatial latents had limitations in the lens of video diffusion models; (1) independent encoding of each frame to a latent limits the consistency across the generated frames, and (2) having one latent for one frame poses a strong limit on the maximum number of frames that can be generated at once, due to high computational overhead.
To this end, recent video generative models operate on the spatio-\textit{temporal} latent space~\cite{hong2022cogvideo, yang2024cogvideox, opensora, opensoraplan, walker2021predicting, kaji2023vq, villegas2022phenaki}, where multiple consecutive frames are encoded into a single latent, and each latent holds not only the appearance information, but also the temporal motion information. 
This approach significantly reduces the number of latents, reducing the computational overhead in both training and inference for diffusion models.

In this work, we propose a new latent space, the spatio-\textit{angular} latent space - encoding both the appearance and the angular viewpoint deviations across multiple viewpoints. 
This is challenging to handle even for existing spatio-temporal latent spaces, as angular viewpoint deviations induces abrupt appearance changes beyond simple motion.
Our proposed spatio-angular latent space enables \methodName to operate on just 8 latents to generate 32 views of an object, exhibiting substantially reduced latency and improved consistency compared to other multi-view diffusion models.

\section{Method: RapidMV}
\label{sec:method}
\begin{figure*}[ht]
    \begin{center}
        \includegraphics[width=1.0\linewidth]{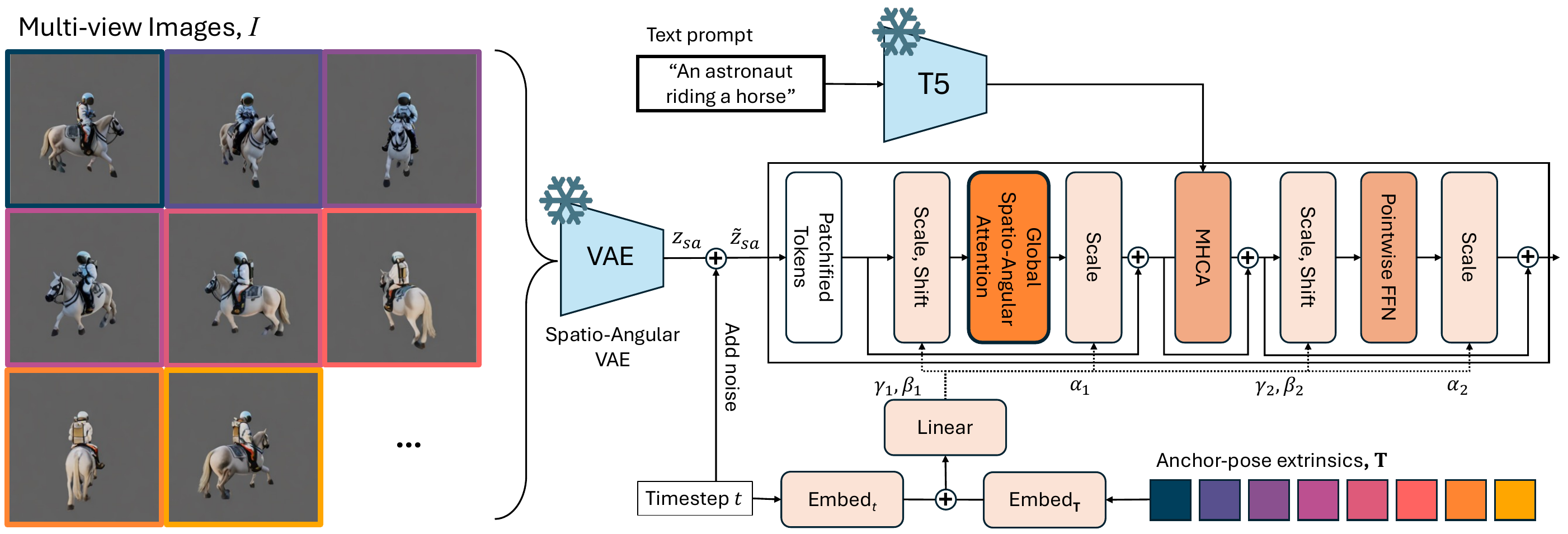}
    \end{center}
    \vspace{-3.0mm}
      \caption{\textbf{Overview of \methodName}. 
      We apply latent-wise anchor-pose modulation, where camera embeddings of the anchor pose corresponding to each spatio-angular latent are added to the timestep embedding to modulate the diffusion blocks, boosting the viewpoint coherency of the generated images.
      We replace the per-frame multi-head self attention with our proposed global spatio-angular attention, facilitating effective communication within and across the spatio-angular latents for enhanced consistency.
}
\label{fig:main_overview}
\end{figure*}

\noindent \textbf{Overview.}
We propose \methodName, a novel multi-view generation model with high quality, consistency and efficiency.
Based on observations from existing multi-view generation or video generation work that using a pretrained text-to-image model is beneficial for the final generation performance~\cite{shi2023mvdream, blattmann2023svd}, we base \methodName on the DiT~\cite{peebles2023scalable}-based text-to-single-view generation model~\cite{chen2024pixartsigma}.
We first introduce our newly proposed \textbf{spatio-angular latent space} (\cref{subsec:spatioangular_vae}), which encodes both the appearance and the angular viewpoint deviation across multiple frames into a single latent for improved efficiency and multi-view consistency. 
We then introduce the overall pipeline of \methodName; 
\methodName facilitates multi-view generation via \textbf{global spatio-angular attention} (\cref{subsec:interlatent_spatioangular_attention}) and \textbf{latent-wise anchor-pose modulation} (\cref{subsec:spatio_angular_modulation}).
For effective training, we decompose our training strategy into progressive steps (\cref{subsec:training_strategy}).
The overall pipeline of our method is illustrated in~\cref{fig:main_overview}.

\subsection{Spatio-angular VAE}
\label{subsec:spatioangular_vae}
\begin{figure}[h]
    \begin{center}
        \includegraphics[width=\linewidth]{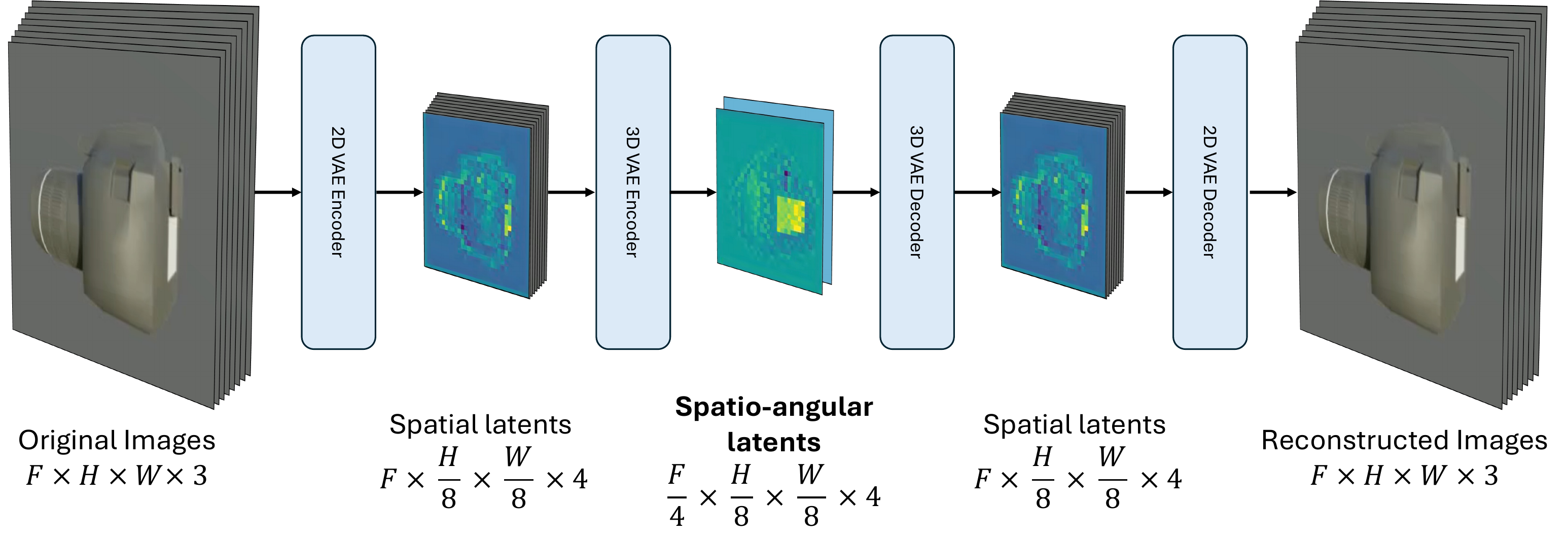}
    \end{center}
    \vspace{-5.0mm}
      \caption{\textbf{Structure of our Spatio-angular VAE.} 
      Spatio-angular VAE yields spatio-angular latents, encoding both appearance and angular viewpoint deviations across adjacent viewpoints into a single latent. \methodName operates in the spatio-angular latent space for improved efficiency and consistency of generation.
}\vspace{-2.0mm}
\label{fig:2d_3d_vae_figure}
\end{figure}

Existing text-to-multi-view diffusion models operate on the spatial latent space; however, we notice that encoding each view into an independent latent in the multi-view setting leads to limited consistency across views, and high computation cost per view. 
To tackle these issues, we newly introduce the spatio-angular latent space, encoding both the appearance and the angular viewpoint deviations across adjacent viewpoints into a single latent.
Inspired by the design of existing spatio-temporal VAEs~\cite{opensora, opensoraplan}, we incorporate existing 2D VAE with 3D VAEs built with causal 3D convolution.
Specifically, we first map each multi-view image $ I^{(i)} \in \mathbb{R}^{H\times W\times 3}$ to independent spatial latents $ z_s^{(i)} \in \mathbb{R}^{\frac{H}{8} \times \frac{W}{8} \times 4}$ using the 2D VAE encoder.
Subsequently, the spatial latents $z_s \in \mathbb{R}^{F \times \frac{H}{8} \times \frac{W}{8} \times 4}$ are compressed along the angular dimension using the 3D VAE encoder to yield a smaller set of spatio-angular latents $z_{st} \in \mathbb{R}^{\frac{F}{4} \times \frac{H}{8} \times \frac{W}{8} \times 4}$. 
In the decoding stage, the spatio-angular latents are first passed to the 3D VAE decoder to reconstruct the spatial latents, which are passed to the 2D VAE encoder to reconstruct the multi-view images $\hat{I}$. 
\cref{fig:2d_3d_vae_figure} illustrates the structure of our spatio-angular VAE.
Assuming we want to generate 32 multi-view images, our spatio-angular latent space allows \methodName to operate on just 8 spatio-angular latents, dramatically improving the efficiency and consistency.

We use a combination of reconstruction loss, LPIPs loss, and KL divergence loss to train our spatio-angular VAE:
\begin{align}
\mathcal{L}_{\text{VAE}} =\ & \frac{1}{N} \sum_{i=1}^N \left( \frac{\| I - \hat{I}_i \|_1}{\exp(\text{logvar})} + \text{logvar} \right) \nonumber \\
& + \lambda_{\text{LPIPS}} \cdot \text{LPIPS}(I, \hat{I}) \nonumber \\
& + \lambda_{\text{KL}} \cdot D_{\text{KL}}\bigl(q(z_{sa} \mid I) \parallel p(z_{sa})\bigr)
\label{eqn:vae_finetune}
\end{align}

\subsection{Global Spatio-angular Attention.}
\label{subsec:interlatent_spatioangular_attention}
The self-attention blocks in text-to-single-view DiT do not attend to patches across different frames; this may suffice for single-view generation, but it is essential to attend across multiple views for consistent and high-quality multi-view generation.
Also, the original self-attention blocks operate on spatial latents, while \methodName handles spatio-angular latents with both appearance and angular viewpoint deviation information.
To facilitate the attention across multiple spatio-angular latents, we propose the \textbf{global spatio-angular attention}, replacing the multi-head self-attention in the PixArt blocks.
Specifically, we can assume a \textit{patchified} noisy latent input $\tilde{z}_{sa}
 \in \mathbbm{R}^{(B' \times \frac{F}{4}) \times T \times d}$, where $T = \frac{H}{8p} \times \frac{W}{8p}$, with $p$ being the patch size.
In the multi-view training of \methodName, we rearrange the patchified latent so that our new attention module can attend to all the spatio-angular latents pertaining to a text prompt, \ie $\tilde{z}^{\textrm{multi}}_{sa} \in \mathbbm{R}^{B' \times (\frac{F}{4} \times T) \times d}$.
Unlike the frame-wise self-attention which operates over $T$ tokens pertaining to a single spatial latent, our inter-latent spatio-angular attention operates over $\frac{F}{4} \times T$ tokens so that each token can attend to the entire set of patchified spatio-angular latents, enhancing the consistency and coherency between the multi-view tokens. 

\subsection{Latent-wise Anchor-pose modulation.}
\label{subsec:spatio_angular_modulation}
One issue with our global spatio-angular attention is that once we reshape the tokens to $\tilde{z}^{\textrm{multi}}_{sa}$, it is hard to discern which tokens are from the same latent.
This is a necessary capability for \methodName, as each spatio-angular latent should be responsible for generating non-overlapping orbital views of an object.
Existing methods~\cite{shi2023mvdream,zuo2024videomv} propose to provide frame-wise camera poses as the condition to the diffusion model; however, this scheme is not straightforward to be applied to spatio-angular latents, as each latent is responsible for multiple adjacent views.
To this end, we propose \textbf{Latent-wise anchor-pose modulation}, \ie we modulate the spatio-angular latents using distinct anchor poses.
Assuming a spatio-angular latent $z_{sa}$ which was obtained from 4 spatial latents $z_s^{i:i+4}$, we declare the $i$-th camera pose as the \textit{anchor} pose of the spatio-angular latent, and use this camera pose to modulate the spatio-angular latent.
The $4\times4$ anchor camera matrix $\mathbf{T}_i$ is flattened, and is embedded using an MLP, \ie $e^i_\mathbf{T} = \mathcal{M}_{\mathbf{T}}(\textrm{vec}(\mathbf{T}^i))$.
The anchor camera embeddings $e_\mathbf{T}^i$ are added to the timestep embeddings $e_t$ in the AdaLN-single layers, to be used to calculate the shifting and scaling terms $\gamma$, $\beta$ and $\alpha$ in the transformer blocks of \methodName. 
As a result, while a single timestep is used to modulate all the spatio-angular latents within a batch, each spatio-angular latent is further modulated by their respective anchor camera poses, facilitating the generation of non-overlapping orbital views.

\subsection{Training strategy decomposition.}
\label{subsec:training_strategy}
Motivated by the efficacy of decomposed training strategy in text-to-image~\cite{chen2023pixartalpha, chen2024pixartsigma} and text-to-video~\cite{hong2022cogvideo, opensora, blattmann2023svd} generation, we propose to decompose our training strategy for improved performance and efficiency of training:
\begin{itemize}
\item We first train \methodName to generate 4 views using \textit{spatial} latents (RapidMV$_s$), where the viewpoints are placed at random azimuthal positions among 32 viewpoints along a horizontally circular orbit. This helps the model to adhere to the camera pose modulation.
\item We then train RapidMV$_s$ to generate 32 views using \textit{spatial} latents, which are placed on even azimuthal distances from each other on a horizontally circular orbit. This stage trains \methodName to be able to generate consistent and dense multi-view images.
\item Based on the 32-view generation model, we replace the spatial VAE with our new \textit{spatio-angular} VAE to leverage spatio-angular latents for efficiency and consistency.
\item We finally finetune \methodName on the high-quality subset of multi-view dataset~\cite{deitke2023objaverse} to enhance the quality and consistency of generated multi-view images.
\end{itemize}

Across all training stages, we inherit PixArt~\cite{chen2024pixartsigma}'s loss formulation to optimize \methodName:
\begin{align}
\mathcal{L}_{\text{MSE}} &= \mathbb{E}_{x_0, \epsilon, t} \left[ \| \epsilon - \hat{\epsilon} \|_2^2 \right], \\
\mathcal{L}_{\text{VB}} &= \mathbb{E}_{x_0, \epsilon, t} \left[ \text{VB}(x_t, x_0, t) \right], \\
\mathcal{L}_{\text{total}} &= \mathcal{L}_{\text{MSE}} + \mathcal{L}_{\text{VB}}
\end{align}
where $\mathcal{L}_{\text{MSE}}$ penalizes incorrect noise predictions, and $\mathcal{L}_{\text{VB}}$ penalizes the variational bound.
\begin{table*}[t]
\centering
\resizebox{\textwidth}{!}{
\begin{tabular}{l ccc cc ccc c}

\toprule

\multirow{2}{*}{Method} & \multicolumn{3}{c}{Quality} & \multicolumn{2}{c}{Text-Image Alignment} & \multicolumn{3}{c}{Consistency} & \multirow{2}{*}{Latency (s)$\downarrow$} \\
\cmidrule(r){2-4}
\cmidrule(r){5-6}
\cmidrule(r){7-9}
& FID$_\textrm{Objaverse}\downarrow$ & FID$_{\textrm{PixArt-}\Sigma}\downarrow$ & IS$\uparrow$ & CLIP-R$\uparrow$ & CLIP Score$\uparrow$ & PSNR$\uparrow$ & LPIPS$\downarrow$ & SSIM$\uparrow$ & \\

\midrule
\multicolumn{10}{l}{\textit{Dataset}} \\
\midrule
Objaverse~\cite{deitke2023objaverse} & - & 96.14 & 17.67$\pm$0.38 & 31.5 & 31.4$\pm$3.25 & 28.14 & 0.1879 & 0.9037 & - \\

\midrule
\multicolumn{10}{l}{\textit{Text-to-Image/Video}} \\
\midrule
PixArt-$\Sigma$~\cite{chen2024pixartsigma} & 96.14 & - & 29.8$\pm$1.17 & 90.7 & 32.8$\pm$2.53 & - & - & - & 0.6 \\
SDXL$_{1024}$~\cite{podell2023sdxl} & 93.87 & 38.51 & 32.0$\pm$1.32 & 91.3 & 33.1$\pm$1.32 & - & - & - & 5.6 \\
OpenSora~\cite{opensora} & 144.37 & 93.45 & 23.2$\pm$0.79 & 85.0 &32.2$\pm$2.73 & - & - & - & 5.7 \\

\midrule
\multicolumn{10}{l}{\textit{Text-to-\textbf{4}-view}} \\
\midrule
MVDream~\cite{shi2023mvdream} & 90.27 & 92.37 & 13.2$\pm$0.99 & 73.3 & \textbf{34.3$\pm$3.36} & - & - & - & 3.6 \\
RapidMV$_s$ (Ours) & \textbf{88.63} & \textbf{89.82} & \textbf{15.6$\pm$1.45} & \textbf{88.0} & 32.6$\pm$2.82 & - & - & - & \textbf{2.7} \\

\midrule
\multicolumn{10}{l}{\textit{Text-to-\textbf{24}-view}} \\
\midrule

MVDream~\cite{shi2023mvdream} & 78.53 & \textbf{76.29} & \textbf{19.4$\pm$0.58} & \textbf{78.1} & \textbf{35.0$\pm$2.95} & 20.05 & \underline{0.4610} & \underline{0.7433} & 34.0 \\
VideoMV$_\textrm{base}$~\cite{zuo2024videomv} & 70.82 & 78.48 & 17.2$\pm$0.57 & 68.9 & 33.6$\pm$2.76 & 19.66 & 0.5597 & 0.6248 & \underline{32.9} \\
VideoMV$_\textrm{gs}$~\cite{zuo2024videomv} & \underline{68.35} & 78.74 & \underline{17.3$\pm$0.70} & 69.0 & \underline{33.8$\pm$2.83} & \underline{21.81} & 0.5134 & 0.7076 & 67.8 \\
\methodName (Ours) & \textbf{60.07} & \underline{77.69} & 16.5$\pm$0.84 & \underline{76.7} & 30.7$\pm$3.10 & \textbf{22.53} & \textbf{0.4149} & \textbf{0.7808}  & \textbf{3.9} \\

\midrule
\multicolumn{10}{l}{\textit{Text-to-\textbf{32}-view}} \\
\midrule

MVDream~\cite{shi2023mvdream} & 77.73 & \textbf{75.08} & \textbf{19.4$\pm$0.64} & \textbf{77.2} & \textbf{34.7$\pm$3.05} & 20.17 & 0.4602 & \underline{0.7461} & 55.9 \\
VideoMV$_\textrm{base}$~\cite{zuo2024videomv} & 71.77 & 78.52 & \underline{19.0$\pm$0.81} & 67.7 & 33.4$\pm$2.81 & 20.18 & 0.5753 & 0.6516 & 41.8 \\
VideoMV$_\textrm{gs}$~\cite{zuo2024videomv} & 68.35 & 78.74 & 18.2$\pm$0.73 & 67.3 & \underline{33.6$\pm$2.78} & \underline{22.29} & 0.5278 & 0.7305 & 86.3 \\
OpenSora$_\textrm{finetuned}$~\cite{opensora} & \underline{67.23}& 101.03 & 11.9±0.30 & 54.0 &27.8$\pm$4.14 & 19.24 & \underline{0.3586} & 0.7053 & \underline{5.7} \\
\methodName (Ours) & \textbf{59.62} & \underline{77.18} & 16.7$\pm$0.84 & \underline{76.6} & 30.7$\pm$3.11 & \textbf{23.01} & \textbf{0.2983} & \textbf{0.8327}  & \textbf{5.3} \\

\bottomrule

\end{tabular}
}
\caption{\textbf{Quantitative evaluation on text-to-multi-view generation.} All resolutions are at $256\times256$ except for SDXL~\cite{podell2023sdxl}, which is at $1024\times1024$. 
RapidMV$_s$ denotes our method which uses the spatial latent space, as 4 orthogonal views is too small to be represented in a spatio-angular space. 
VideoMV$_\textrm{base}$ are results of VideoMV~\cite{zuo2024videomv} before the 3D-aware denoise sampling, and VideoMV$_\textrm{gs}$ are results of VideoMV after the 3D-aware denoise sampling, consequently exhibiting higher latency.
Across all number of views, \methodName demonstrates the best FID$_\textrm{Objaverse}$, consistency and latency, while being competent on FID$_{\textrm{PixArt-}\Sigma}$ and CLIP-R scores.
Notably, \methodName can generate 32 views in just around 5 seconds, shorter than the time taken for SDXL to generate a single $1024\times1024$ image. and $\sim8$ times faster than the previously fastest method, VideoMV$_\textrm{base}$.
}
\vspace{-1.0mm}
\label{tab:4-view-quantitative}
\end{table*}

\begin{figure*}[ht]
    \begin{center}
        \includegraphics[width=\linewidth]{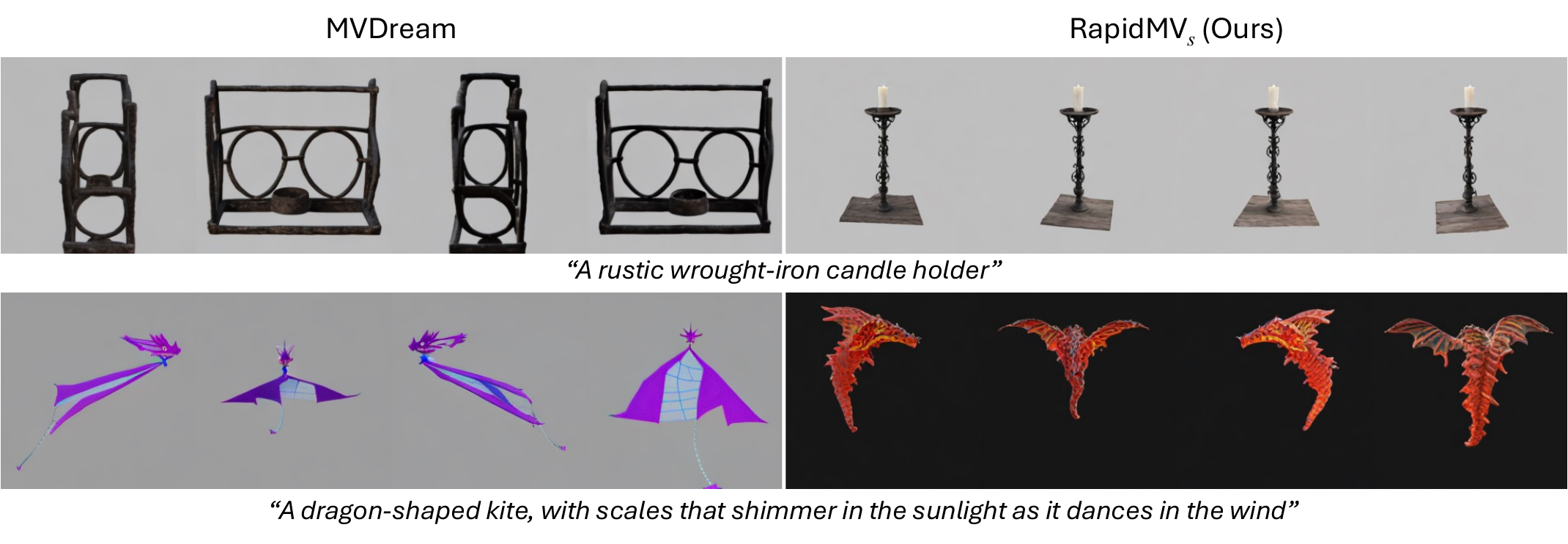}
    \end{center}
    \vspace{-5.0mm}
      \caption{\textbf{Text-to-4-view results compared against MVDream~\cite{shi2023mvdream}}. 
      It can be seen that RapidMV$_s$ yields high-quality, consistent 4-view images which show strong text-image alignment compared to MVDream.
}
\label{fig:main_qual_4view}
\end{figure*}

\begin{figure*}[ht]
    \begin{center}
        \includegraphics[width=\linewidth]{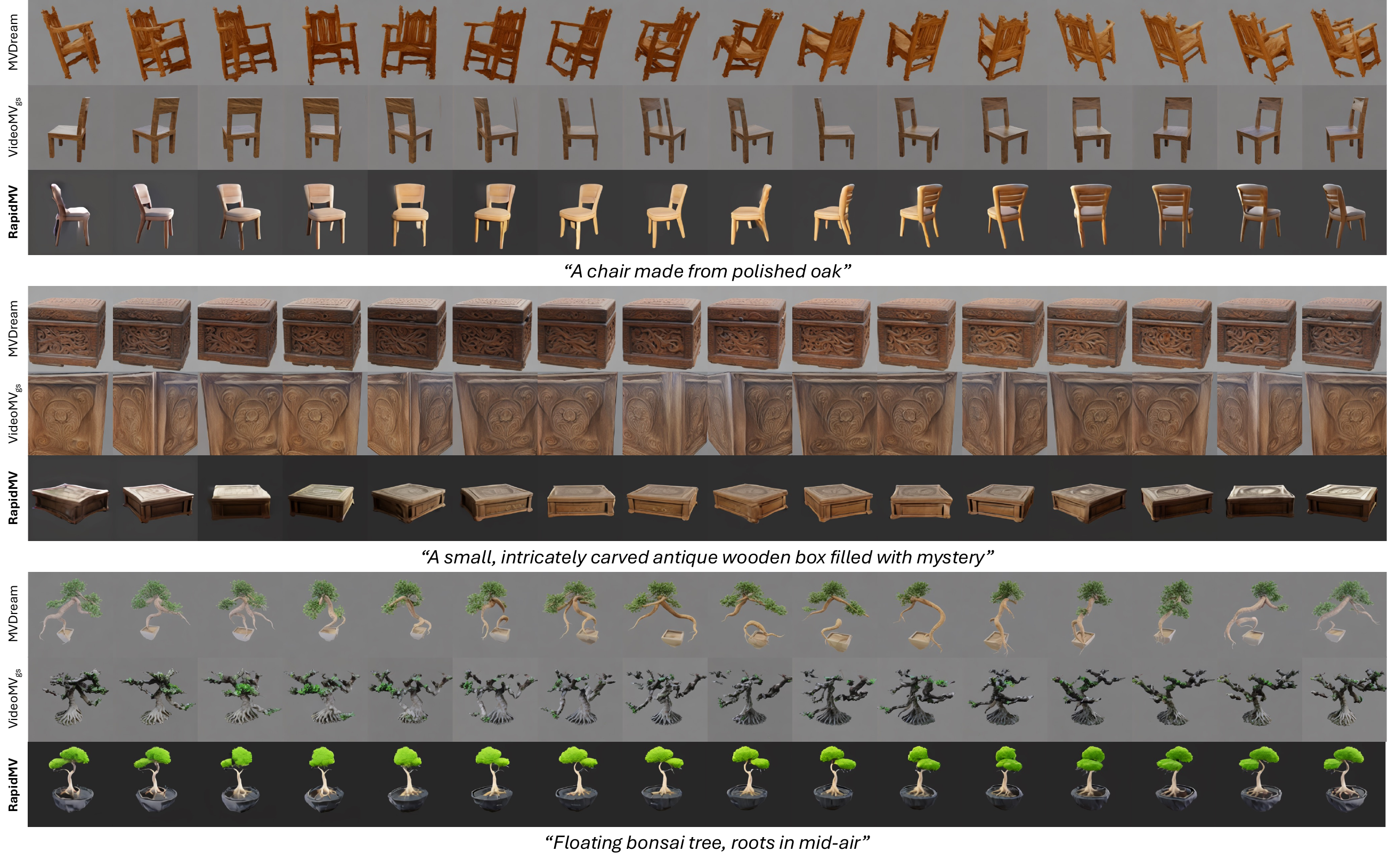}
    \end{center}
    \vspace{-5.0mm}
      \caption{\textbf{Text-to-32-view results of \methodName in comparison to MVDream~\cite{shi2023mvdream} and VideoMV~\cite{zuo2024videomv}}.
    We visualize 16 even number frames in this figure for better visibility.
      It can be seen that \methodName yields favorable quality and consistency, and also exhibit strong text-image alignment.
      Best viewed on electronics.
}
\label{fig:main_qual_32view}
\end{figure*}

\section{Experiment}
\label{sec:experiment}

\begin{table*}[t]
\centering
\resizebox{\textwidth}{!}{
\begin{tabular}{l ccccc cccc}

\toprule

\multirow{2}{*}{Method} & \multicolumn{3}{c}{Quality} & \multicolumn{2}{c}{Text-Image Alignment} & \multicolumn{3}{c}{Consistency} & \multirow{2}{*}{Latency (s)$\downarrow$} \\
\cmidrule(r){2-4}
\cmidrule(r){5-6}
\cmidrule(r){7-9}
& FID$_\textrm{Objaverse}\downarrow$ & FID$_{\textrm{PixArt-}\Sigma}\downarrow$ & IS$\uparrow$ & CLIP-R$\uparrow$ & CLIP Score$\uparrow$ & PSNR$\uparrow$ & LPIPS$\downarrow$ & SSIM$\uparrow$\\
\midrule
\multicolumn{9}{l}{\textit{Text-to-\textbf{32}-view}} \\
\midrule
RapidMV$_s$ & 68.51 & \textbf{75.22} & \textbf{18.7$\pm$0.57} & 74.8 & \underline{30.5$\pm$3.28} & 21.19 & \underline{0.3381} & 0.7718 & \underline{30.7} \\
RapidMV$_{st}$ & 66.59 & 95.32 & 14.1$\pm$0.42 & 62.8 & 28.7$\pm$3.80 & 21.15 & 0.3892 & 0.7628 & \textbf{5.3} \\
\methodName (Ours) & \textbf{{59.62}} & \underline{77.18} & \underline{16.7$\pm$0.84} & \textbf{76.6} & \textbf{30.7$\pm$3.11} & \textbf{23.01} & \textbf{0.2983} & \textbf{0.8327} & \textbf{5.3}\\
- w/o RapidMV$_s$ pretrain & 64.52 & 89.15 & 14.2$\pm$0.81 & 64.6 & 29.5$\pm$3.42 & 21.72 & 0.3546 & 0.7843 & \textbf{5.3} \\
- w/o hq finetuning & \underline{60.79} & 82.81 & 14.7$\pm$0.78 & \underline{75.9} & 30.4$\pm$3.17 & \underline{21.94} & 0.3477 & \underline{0.7986} & \textbf{5.3} \\
\bottomrule

\end{tabular}
}
\caption{\textbf{Comparative evaluation for design choices of \methodName.}
The results show that using the spatio-angular space (\methodName) shows the best results overall, in comparison to using the spatial latents (RapidMV$-s$) or spatio-temporal latents (RapidMV$_{st})$.
While RapidMV$_s$ does show improved FID$_{\textrm{PixArt}\Sigma}$ and IS, RapidMV exhibits $\sim6$ times lower latency in comparison. 
We also show that building RapidMV on top of RapidMV$_s$ as part of our training scheme leads to improved results, and that our final high-quality finetuning stage yields overall improvementes in terms of quality, text-image alignment, and consistency.
}
\vspace{-1.0mm}
\label{tab:main_ablation}
\end{table*}

\begin{figure}[h]
    \begin{center}
        \includegraphics[width=\linewidth]{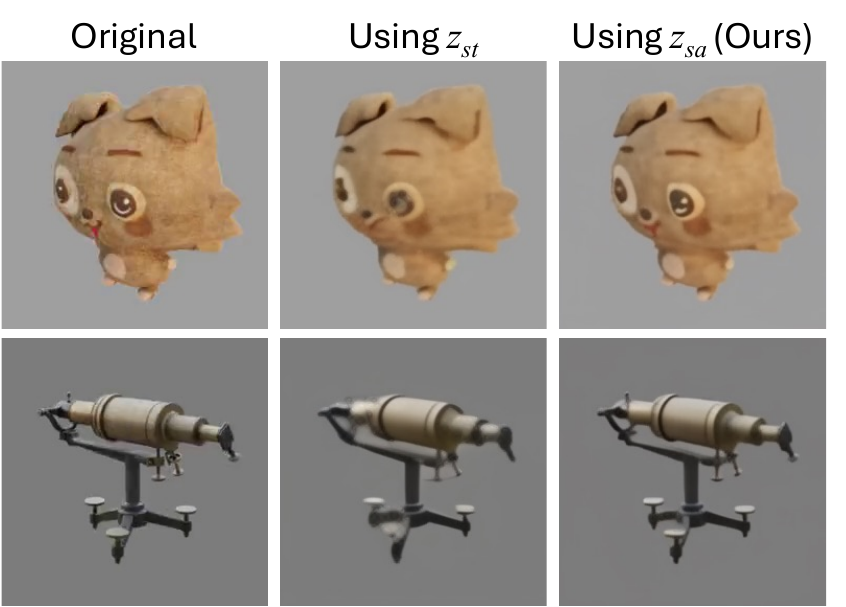}
    \end{center}
    \vspace{-5.0mm}
      \caption{\textbf{Reconstruction quality comparison of off-the-shelf spatio-temporal latents v.s. our proposed spatio-angular latents.}
    Compared to using the off-the-shelf spatio-temporal latents from the OpenSora-VAE v1.2~\cite{opensora}, we show that our newly proposed spatio-angular latents yields better reconstruction results, better preserving the details.
      We conjecture this is because the orbital views we are trying to create has high motion dynamics, which is not common in many video datasets which were used to train the spatio-temporal VAE.
}
\label{fig:ablation_vae_finetuning}
\end{figure}

\begin{figure}[t]
    \begin{center}
        \includegraphics[width=\linewidth]{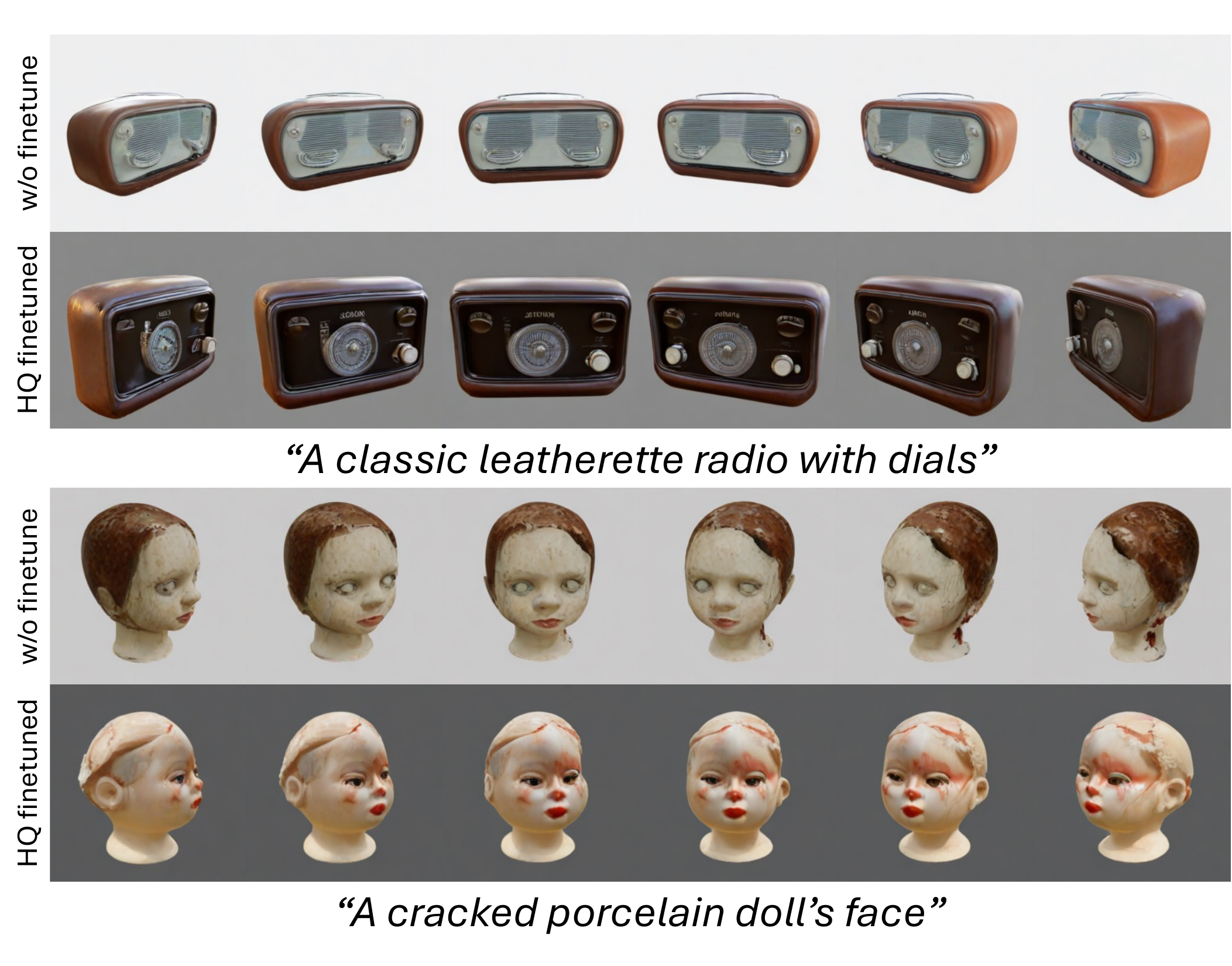}
    \end{center}
    \vspace{-6.0mm}
      \caption{\textbf{Effect of high-quality finetuning of \methodName}. 
      High-quality finetuning improves the generation quality of \methodName.
}
\label{fig:ablation_hq_finetuning}
\end{figure}

\subsection{Training dataset}
\label{subsec:training_dataset}
We render multi-view images from the Objaverse~\cite{deitke2023objaverse} dataset to use as our training dataset.
We follow the rendering protocols used in MVDream~\cite{shi2023mvdream}, finally obtaining rendered views of approximately 350K distinct objects. 
Observing that PixArt~\cite{chen2023pixartalpha, chen2024pixartsigma} facilitated effective and efficient training via the use of densely captioned images, we also use the dense captions provided by BS-Objaverse 660K introduced in Bootstrap 3D~\cite{sun2024bootstrap3d}.
For overlapping objects between our dataset and BS-Objaverse 660K ($\sim$250K objects), we use dense captions; otherwise, we use the names and tags of the objects provided by the Objaverse dataset as the text caption.
When constructing the high-quality subset of Objaverse, we filter out objects with less than 10 `likes' in the metadata; we visually confirmed that the overall complexity and quality of the 3D objects in Objaverse are strongly correlated with the number of likes, as demonstrated in~\cref{fig:supp_hq_viz} of the supplementary. 
This yields around 70k high-quality objaverse data.

Motivated by the fact that many multi-view diffusion models yielded improved quality by mixing 2D image data in their training~\cite{shi2023mvdream, sun2024bootstrap3d, zuo2024videomv}, we also incorporate a subset ($\sim$500K) of SA-1B dataset used in training SAM~\cite{kirillov2023segment}, adhering to PixArt~\cite{chen2023pixartalpha, chen2024pixartsigma}.
We use SAM-LLaVA-Captions10M provided by PixArt, which contains dense captions with high concept density for improved text-image alignment. 
During training, we train on the multi-view images of Objaverse for 70\% of the time, and the single-view images of SAM for 30\% of the time.

\subsection{Implementation details}
\textbf{\methodName.}
\methodName was initialized from the pretrained PixArt-$\Sigma$-512~\cite{chen2024pixartsigma} model to benefit from its text-to-image generation capabilities and scalability.
We train \methodName to generate multi-view images at image dimensions of $256 \times 256$.
\methodName assumes that the object described by the text prompt is placed in the center and the camera is orbiting around the object at a fixed elevation.
All stages of training are run on 8 A100 80GB GPUs, with a batch size (total number of images) of 128 per GPU \ie 4 objects within a batch when $F=32$.
We trained \methodName through each stage of training for 50K steps, except for the final high-quality finetuning stage, which we trained for 20K steps.  
We optimize \methodName using the AdamW~\cite{loshchilov2017decoupled} optimizer, at a constant learning rate of $1e^{-5}$.

As mentioned above, we train \methodName on single-view images for 30\% of the time; to make training with single-view images compatiable with our spatio-angular latent space, we concatenate each single-view image with its copy three times, and encode the resulting set of four identical images into a single spatio-angular latent for training. 
When training with images, we perform local spatio-angular attention instead of global spatio-angular attention \ie each spatio-angular latent attends to only itself, as different 2D images do not need to attend to one another for generation.
Also, we disable latent-wise anchor-pose modulation when training with single-view images, as single-view images do not have a predefined camera pose. 

During inference, we use DPMSolver++~\cite{lu2022dpm} for sampling with 14 steps, and a CFG guidance scale of $6.0$. 

We base our spatio-angular VAE on the spatio-temporal VAE from OpenSora v1.2~\cite{opensora}, to benefit from the large-scale video pretraining of OpenSora's VAE.
To train our VAE to map orbital views of an object to the spatio-angular latent space, we use the 32-orbital view renderings of the Objaverse dataset as the finetuning dataset. 
We use the loss function illustrated in ~\cref{eqn:vae_finetune} to optimize our VAE to successfully reconstruct the orbital multi-view images.
In training our spatio-angular VAE, we use a batch size of 1 ($F=32$), and use the Adam~\cite{kingma2014adam} optimizer with a learning rate of $1e^{-5}$.
We set $\lambda_\textrm{LPIPS}=0.1$, and $\lambda_\textrm{KL}=1e^{-6}$.

\subsection{Evaluation of Text-to-Multiview generation.}
\label{subsec:t2mv_eval}

\subsubsection{Qualitative \& quantitative comparison}

\noindent
\textbf{Dataset \& baselines.}
We evaluate \methodName on 210 distinct prompts; 100 single-object prompts of T3bench~\cite{he2023t}, and 110 prompts from GPTEval3D~\cite{wu2024gpt}.
We evaluate the quality, image-text alignment, multi-view consistency, and the latency of \methodName against the baseline methods of MVDream~\cite{shi2023mvdream} and VideoMV~\cite{zuo2024videomv}.
MVDream proposes to generate 4 orthogonal views, while VideoMV proposes to generate 24 views; to this end, we make comparisons across 4, 24 and 32 views, using our spatial-latent version for 4-view comparison.
We also finetune OpenSora on 32-frame orbital videos of Objaverse for a fair comparison (OpenSora$_\textrm{finetuned}$) against spatio-temporal latents.

\smallbreak
\noindent
\textbf{Evaluation metrics.}
For quality evaluation, we report (1) FID$_\textrm{Objaverse}$: the FID against 32 views of 1000 high-quality Objaverse objects (32k images),  (2) FID$_{\textrm{PixArt-}\Sigma}$: the FID against 32 randomly generated images by PixArt-$\Sigma$ for each of the 210 prompts, and (3) the inception score (IS).
For text-image alignment, we report (1) the CLIP-R score, which measures the recall based on the CLIP feature similarity, and (2) the CLIP score, which quantifies the similarity between the image and text CLIP features.
For consistency evaluation, given $F$ generated views, we use all even-numbered frames to optimize a NeRF using the nerfacc~\cite{li2023nerfacc} implementation of Instant-NGP~\cite{muller2022instant}.
We then render the odd-numbered views from the optimized NeRF, and measure the PSNR, LPIPS and SSIM as metrics for consistency.
We do not measure the consistency for 4-view generation, since 4 views are insufficient for reconstruction.
The results are illustrated in~\cref{tab:4-view-quantitative}.

\smallbreak
\noindent
\textbf{Results \& discussion.}
We show that across 4, 24 and 32 views, \methodName exhibit the best FID$_\textrm{Objaverse}$ and consistency (PSNR, LPIPS, SSIM).
\methodName also shows the lowest latency, demonstrating its superior efficiency in generating multi-view images. 
\methodName shows competitive FID$_{\textrm{PixArt-}\Sigma}$ and CLIP-R scores, while MVDream shows the best CLIP Score.
This shows that \methodName achieves competitive image quality, while outperforming existing methods in terms of multi-view consistency and efficiency.
We provide qualitative results for 4-view generation of RapidMV$_s$ in~\cref{fig:main_qual_4view}, and 32-view generation of RapidMV in~\cref{fig:main_qual_32view}, where it can be seen that \methodName yields visually compelling results with high quality and consistency. 

\subsubsection{User study}
We perform a user study using 30 randomly selected prompts from GPTEval3D.
For each of the prompts, we render 32-view orbital images, and render them into a rotating video.
Each user is asked to select the preferred orbital video in terms of the quality and consistency.
We collect $1200$ responses and report them below in~\cref{tbl:rebuttal_user_study}; it can be seen that our proposed \methodName shows the highest user preference. 

\begin{table}[h]
    \centering
    \scalebox{1.0}{
       \begin{tabular}{lc}
                \toprule

                Method & User preference \% \\
                
                \midrule

                MVDream$_\textrm{finetuned}$ & 25.2 \\
                VideoMV & 31.5 \\
                \methodName & 43.3 \\ 
                \bottomrule
        \end{tabular}    
         }
        \vspace{-1.5mm}
\caption{\textbf{User study on 32-view generation.}
\methodName shows the highest user preference on quality and consistency.
}
\vspace{-3.00mm}
\label{tbl:rebuttal_user_study}
\end{table}

\subsection{Comparative analyses}
We evaluate the (1) comparative abilities of spatial spatio-temporal, and spatio-angular latents, (2) the efficacy of our decomposed training strategy, and (3) the effect of finetuning \methodName with high-quality data in~\cref{tab:main_ablation}.
The results show that using our proposed spatio-angular latent space yields the best overall results, in terms of quality, text-image alignment, consistency, and latency.
While using the spatial latents (RapidMV$_s$) yields strong generation quality as well, the latency of using spatio-angular latents is $\sim6\times$ faster than using spatial latents, with improved multi-view consistency as well.
We visualize the comparative reconstruction results using spatio-temporal and spatio-angular latent space in~\cref{fig:ablation_vae_finetuning}, where it can be seen that using spatio-temporal latents to reconstruct orbital videos results in poor reconstruction quality.
The results also show that our decomposed training scheme leads to progressively improved results.
We additionally visualize the generated multi-view images from \methodName with and without high-quality fine-tuning in~\cref{fig:ablation_hq_finetuning}, where it can be seen that high-quality finetuning leads to noticeable quality improvements in the final generated results.

\section{Conclusion}
\label{conclusion}

In this work, we present \methodName, an efficient and effective text-to-multiview diffusion model that generates dense, consistent multi-view images.
In essence, we leverage spatio-angular latents to efficiently encode both appearance and angular viewpoint deviations into a single latent, dramatically improving efficiency and consistency of \methodName.
\methodName builds on a DiT-based text-to-single-view diffusion framework by (1) applying global spatio-angular attention across all spatio-angular latents to enhance multi-view consistency, and (2) integrating extrinsic camera control via latent-wise anchor-pose modulation, ensuring camera pose coherence across views.
Our quantitative and qualitative evaluations demonstrate that \methodName efficiently produces high-quality, multi-view images with state-of-the-art consistency, and that our proposed training scheme decomposition leads to progressive improvements in performance in comparison to direct training.
\methodName alleviates the issues of low density and consistency in existing text-to-multi-view generation methods, paving the way to effective and efficient 3D generation.

\clearpage
\appendix
\setcounter{table}{0}
\renewcommand{\thetable}{A\arabic{table}}%
\setcounter{figure}{0}
\renewcommand{\thefigure}{A\arabic{figure}}%
\maketitlesupplementary

\section{Details of reconstruction evaluation}
\label{appendix:reconstruction_evaluation}

We elaborate on our reconstruction evaluation scheme, which was briefly mentioned in~\cref{subsec:t2mv_eval}.
Given $F$ generated images, where $F \in \{24,32\}$ in our experiments, we use even-numbered frames (half of the generated images) as the train set to train a NeRF, and use the odd-numbered frames (remaining half of the generated images) as the test set. 
This splitting strategy ensures that the training and test sets are mutually exclusive and that the model is evaluated on unseen viewpoints.
For the NeRF, we use the nerfacc~\cite{li2023nerfacc} implementation of Instant-NGP~\cite{muller2022instant} for fast training.
We use multi-resolution hash grids with a resolution of 128 and 4 levels.
We used two separate MLPs to predict volume density and view-dependent color, and we utilized sigmoid for output layers to ensure outputs are in valid ranges.
The learning rate was initialized at 1e-2 with a warm-up period of 100 steps, where the learning rate linearly increased from 1\% to 100\% of its maximum value.
We used a multi-step scheduler that decayed the learning rate by a factor of 0.33 at 50\%, 75\%, and 90\% of the total training steps.
We used weight decay to regularize the model and prevent overfitting, set to 5e-4.
The smooth L1 was used between the predicted and ground-truth RGB values to optimize the NeRF without additional objectives \eg LPIPs loss, at an aim to benchmark the actual photometric multi-view consistency.

One issue we encountered was that we could not pre-determine the camera intrinsics of the generated views, as we apply the latent-wise anchor-pose modulation using the camera extrinsics only.
Our training data rendered from Objaverse~\cite{deitke2023objaverse} had varying intrinsics, as we followed MVDream~\cite{shi2023mvdream} to use a random field of view between $[15^{\circ}, 60^{\circ}]$ for improved diversity in renderings.
While VideoMV~\cite{zuo2024videomv} also reported consistency metrics for MVDream and VideoMV, their reconstruction pipeline was not released for reproduction. 
To resolve this, we devise a scheme to optimize a NeRF for each FoV in $[15^{\circ}, 60^{\circ}]$ at $5^{\circ}$ intervals, iterating for 1,000 steps per configuration.
We then identify the FoV that minimizes the PSNR and select the corresponding NeRF for further optimization with an additional 1,000 steps.
This ensures that each method is evaluated as fairly as possible, leveraging its optimal multi-view consistency for the final comparison.

\begin{figure*}[ht]
    \begin{center}
        \includegraphics[width=1.0\linewidth]{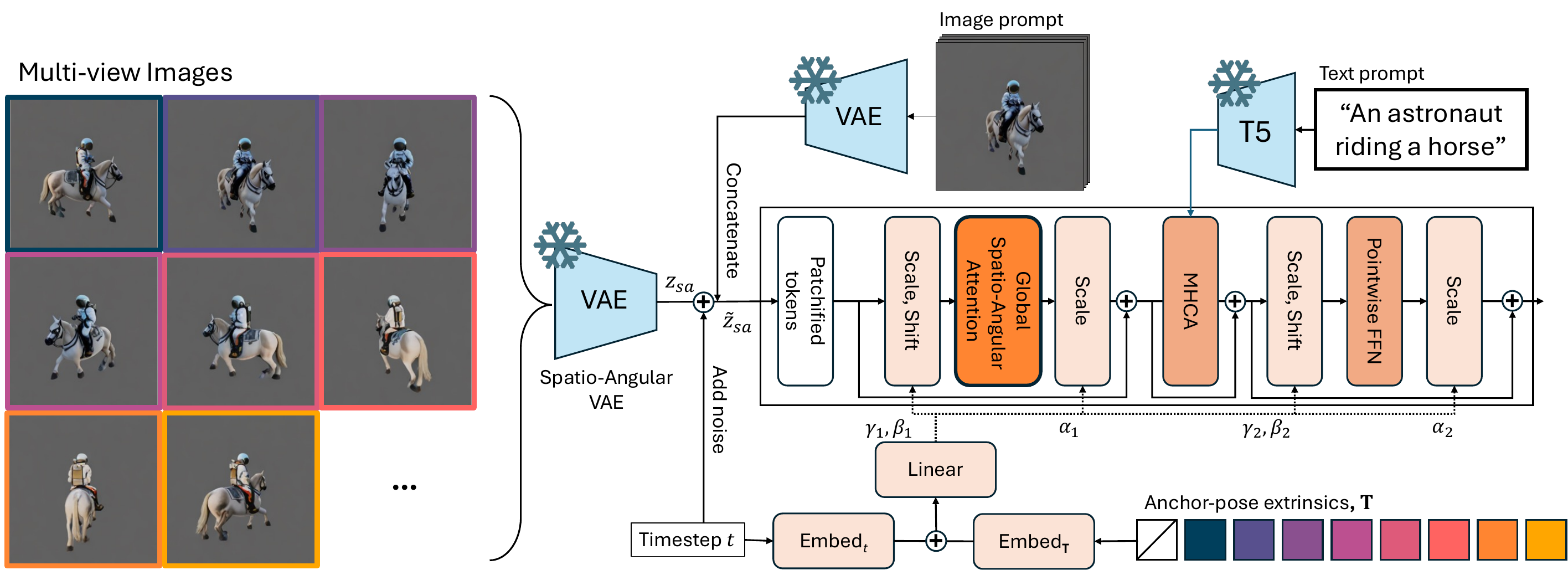}
    \end{center}
    \vspace{-3.0mm}
      \caption{\textbf{Overview of Image-conditioned \methodName}.
      Zero vectors are provided for the latent-wise anchor-pose modulation for the image prompt.
      The latent of the image prompt is concatenated to the \textit{noisy} spatio-angular latents, as the image prompt latent should not be noisy, and is not denoised during the diffusion steps.
}
\vspace{-3.0mm}
\label{fig:supp_image_condition_overview}
\end{figure*}

\begin{figure*}[ht]
    \begin{center}
        \includegraphics[width=1.0\linewidth]{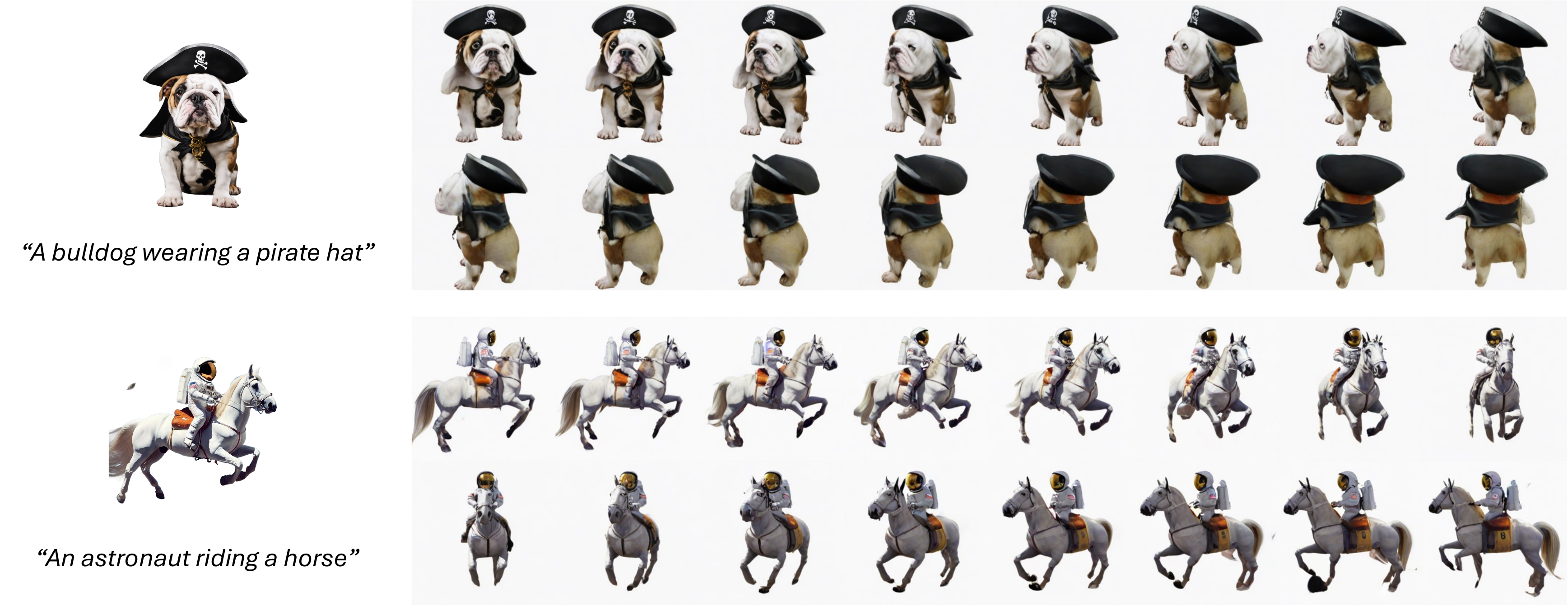}
    \end{center}
    \vspace{-3.0mm}
      \caption{\textbf{Qualitative results of image-conditioned \methodName}. 
      We show that \methodName can flexibly handle image prompts to generate 32 consistent views.
      We visualize 16 contiguous frames in this figure for better visibility.
}
\vspace{-3.0mm}
\label{fig:image_conditioned_mvart}
\end{figure*}

\section{Image-conditioned RapidMV}
\label{appendix:image_conditioning}

In this section, we show that \methodName can be optimized to generate multi-view images conditioned on not only text, but also image. 
The idea is simple; we concatenate the latent of the image prompt to the noisy multi-view latents, so that the multi-view latents can attend to the image prompt during the denoising process for an explicit image guidance. 
The overall implementation is motivated by the pixel controller of ImageDream~\cite{wang2023imagedream, kim2024multiimagedream}.
The frame-wise camera conditioning takes as input a zero vector as the camera extrinsics of the image prompt.
The image prompt latent is not added with noise, and is not denoised during the diffusion process as well.
For image-conditioned \methodName, the image prompt latent is obtained by simply passing four copies of the image prompt through our spatio-angular VAE to obtain a single spatio-angular latent.
The overall pipeline for image-conditioned \methodName is illustrated in~\cref{fig:supp_image_condition_overview}, and we provide qualitative examples in~\cref{fig:image_conditioned_mvart}.

\begin{figure*}[htp]
    \begin{center}
        \includegraphics[width=1.0\linewidth]{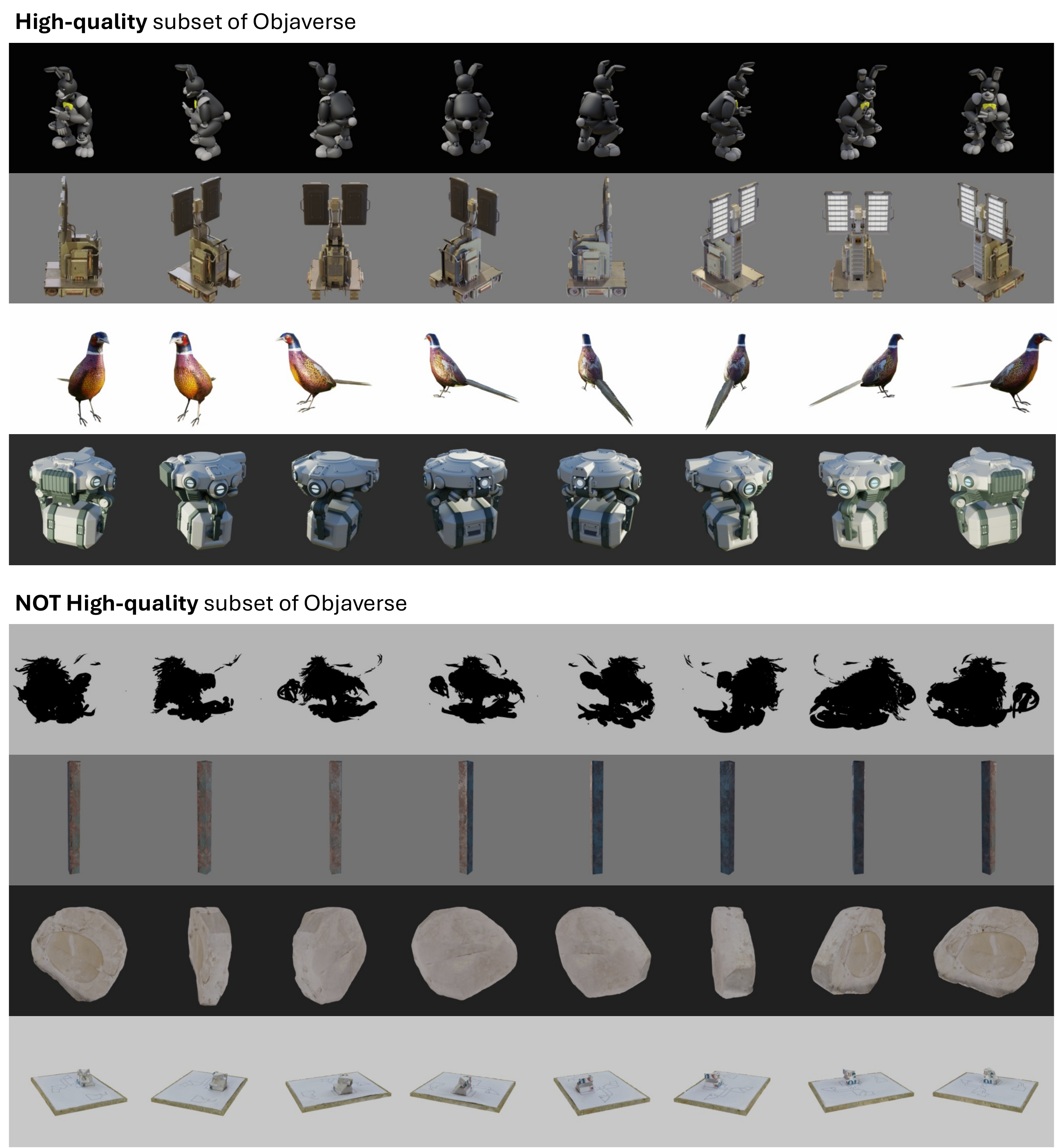}
    \end{center}
      \caption{\textbf{Visualization of high-quality subset of Objaverse~\cite{deitke2023objaverse}.}
      It can be seen that our high-quality subset contains objects with more sophisticated and detailed geometry and texture. 
      We filter out objects from the Objaverse data whose `like' counts in the metadata is less than 10.
      While it is not always the case that objects with lower than 10 likes counts have simple geometry and texture, the like count serves as a reliable metric to yield high-quality objects from the full dataset.
      }

\label{fig:supp_hq_viz}
\end{figure*}

\section{Filtering high-quality data from Objaverse}
\label{appendix:hq_objaverse_visualization}

We mentioned in~\cref{subsec:training_strategy} that we decompose the training strategy, where we finally finetune our model on the high-quality subset of Objaverse~\cite{deitke2023objaverse}.
It was explained in~\cref{subsec:training_dataset} that we filter out objects with less than 10 `likes' in the metadata to collect our high-quality subset, which leaves around 70K objects. 
The efficacy of high-quality finetuning was demonstrated in~\cref{tab:main_ablation} and~\cref{fig:ablation_hq_finetuning}.

In this section, we visualize some examples of our high-quality subset of Objaverse, in contrast to the objects which are not included in our high-quality subset, in~\cref{fig:supp_hq_viz}.
can be seen that our high-quality subset contains objects with more sophisticated and detailed geometry and texture.
While it is not always the case that objects with lower than 10 likes counts have simple geometry and texture, the like count serves as a reliable metric to yield high-quality objects from the full dataset.

\begin{table*}[t]
\centering
\begin{tabular}{l cccc}

\toprule

\multirow{2}{*}{Method} & \multicolumn{2}{c}{CLIP-R}$\uparrow$ & \multicolumn{2}{c}{CLIP Score}$\uparrow$ \\
\cmidrule(r){2-3}
\cmidrule(r){4-5}
& CLIP-L/14 & CLIP-bigG & CLIP-L/14 & CLIP-bigG \\

\midrule
Instant3D~\cite{li2023instant3d}* & 83.6 & 91.1 & 25.6 & 39.2 \\ 
MVDream~\cite{shi2023mvdream} & 84.8 & 89.3 & 25.5 & 38.4 \\
Bootstrap3D~\cite{sun2024bootstrap3d} & \underline{88.8} & \underline{92.5} & \underline{25.8} & 40.1 \\ 
RapidMV$_s$ (ours) & \textbf{90.0} & \textbf{93.4} & \textbf{26.3} & \underline{39.5} \\

\bottomrule

\end{tabular}
\caption{\textbf{Quantitative comparison against Bootstrap3D~\cite{sun2024bootstrap3d} on 4 generated views.}
The evaluation was performed on the 110 prompts from GPTEval3D~\cite{wu2024gpt}.
Instant3D*~\cite{li2023instant3d} are results from an unofficial implementation by the authors of Bootstrap3D.
All resolutions are at $256\times256$. 
The results show that our proposed \methodName exhibits the best CLIP-R score overall, and the best CLIP-Score when using the CLIP-L/14 model~\cite{radford2021learning} and the second-best when using the CLIP-bigG model~\cite{ilharco2022patching}.
}
\label{tab:supp-bootstrap3d}
\end{table*}

\section{Comparison against Bootstrap3D}
\label{appendix:bootstrap3d_comparison}

In this section, we evaluate \methodName against Bootstrap3D~\cite{sun2024bootstrap3d}, a concurrent 4-view genereation model that proposes to use (1) densified captions, (2) large-scale synthetic multi-view dataset and (3) Training-time step Reschedule (TTR) to better leverage the synthetic dataset. 
Their model and pretrained weights were not open-source at the time of submission, and we try to evaluate as fairly as possible by using the same evaluation dataset from GPTEval3D~\cite{wu2024gpt}.
We do not have Bootstrap3D's generated image set from PlayGround2.5 and PixArt-$\alpha$ for FID calculation, and therefore omit the FID value comparisons. 
The results are shown in~\cref{tab:supp-bootstrap3d}, where it can be seen that \methodName outperforms Bootstrap3D in terms of CLIP-Recall, while being competitive in terms of CLIP-score.

\section{Drawbacks and future directions.}
\label{sec:appendix_drawbacks}

A drawback in the current version of \methodName is that it generates multi-view images within a static orbit at fixed elevation.
However, it has been shown in SV3D~\cite{voleti2025sv3d} that having a dynamic orbit, \ie, varying elevation of camera poses covering more various viewpoints, is definitely beneficial in 3D reconstruction.
This could be achieved by rendering views from the Objaverse~\cite{deitke2023objaverse} dataset at dynamic orbits for training, as the camera conditioning scheme would still be applicable to cameras in a dynamic orbit, and spatio-temporal compression would still be effective.

Another shortcoming of \methodName is that even after finetuning, the VAE still is not perfect at alleviating blurry textures or motion blurs, as a compromise for the efficiency of spatio-angular latents.
We hypothesize this is because each latent has to encode not only the appearance of the original image, but also the \textit{angular viewpoint deviations} across 4 frames, which makes it challenging to seamlessly reconstruct the fine details. 
While our current spatio-angular VAE yields a 4-channel spatio-angular latent, more recent spatial VAEs~\cite{esser2024sd3, flux} and spatio-temporal VAEs~\cite{yang2024cogvideox, cvvae} pproduces 16-channel latents, which we conjecture would be more effective at capturing both the appearance and motion information accurately.

The blurring effects are particularly pronounced in the first frame of generation, which we conjecture is due to the \textit{causal} 3D convolution layers within the spatio-angular VAE. 
We conjecture this can be solved if we propose to encode $1+4N$ frames, where the first frame is encoded separately to better preserve the details and to be usable for individual images, following recent spatio-temporal VAE structures~\cite{yang2024cogvideox}.

\begin{figure*}[ht]
    \begin{center}
        \includegraphics[width=\linewidth]{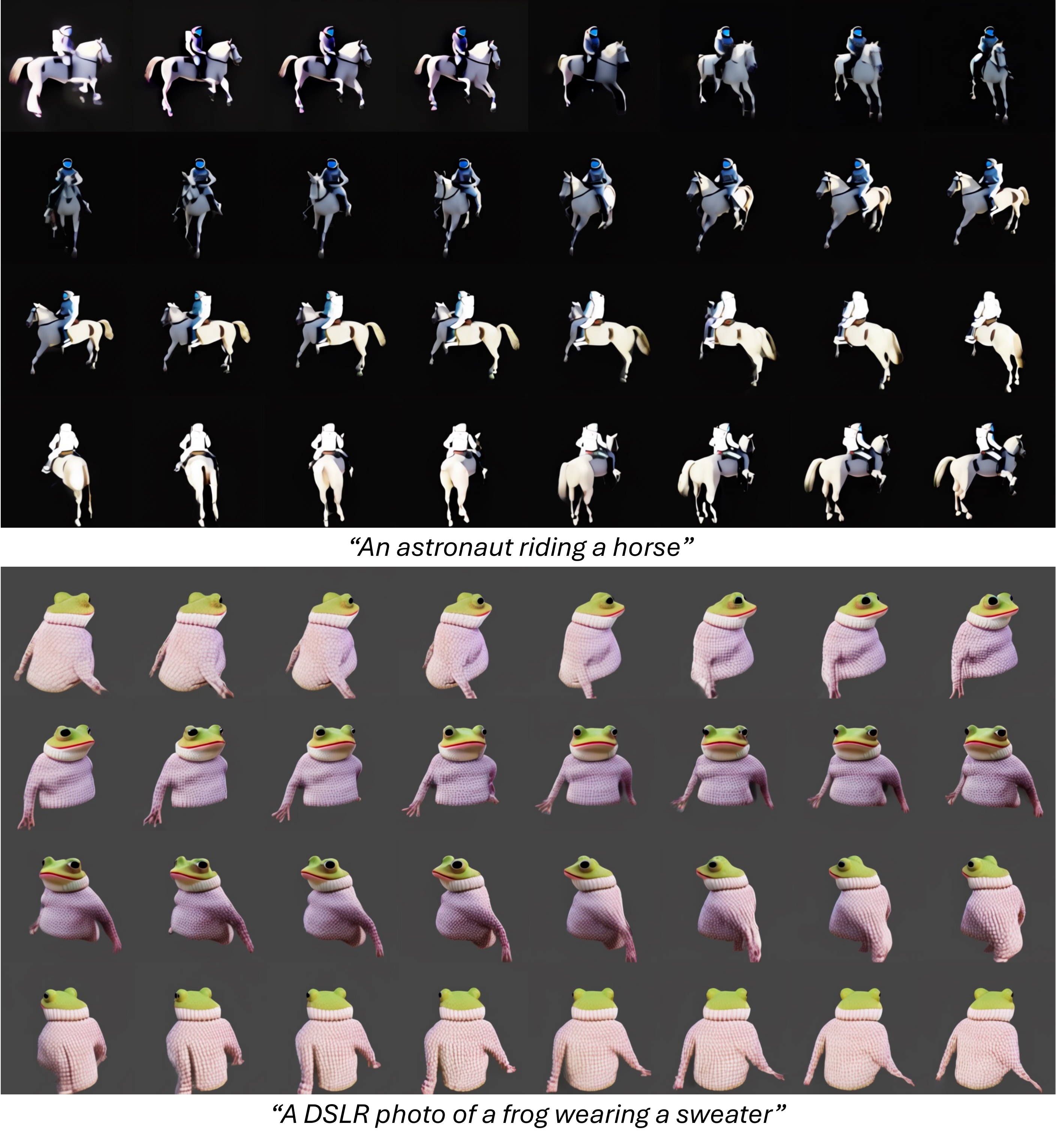}
    \end{center}
      \caption{\textbf{Text-to-32-view results of \methodName}.
    We visualize all 32 frames in this figure for better visibility.
    \methodName shows promising quality and high multi-view consistency and camera coherency, despite generating 32 images in just around 5 seconds.
}
\label{fig:supp_qual_32_b}
\end{figure*}

\begin{figure*}[ht]
    \begin{center}
        \includegraphics[width=\linewidth]{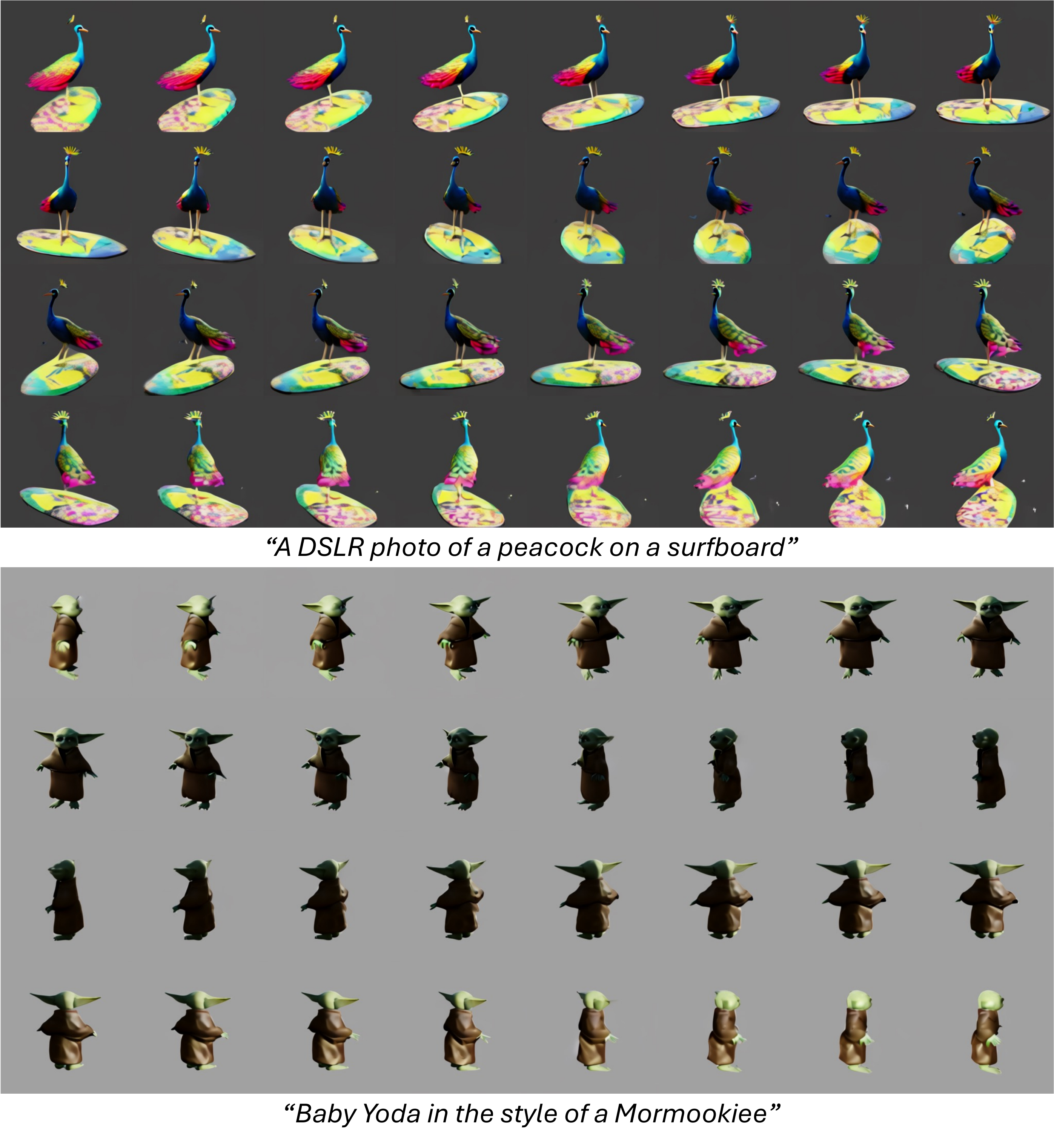}
    \end{center}
      \caption{\textbf{Text-to-32-view results of \methodName}.
    We visualize all 32 frames in this figure for better visibility.
    \methodName shows promising quality and high multi-view consistency and camera coherency, despite generating 32 images in just around 5 seconds.
}
\label{fig:supp_qual_32_c}
\end{figure*}

\begin{figure*}[ht]
    \begin{center}
        \includegraphics[width=\linewidth]{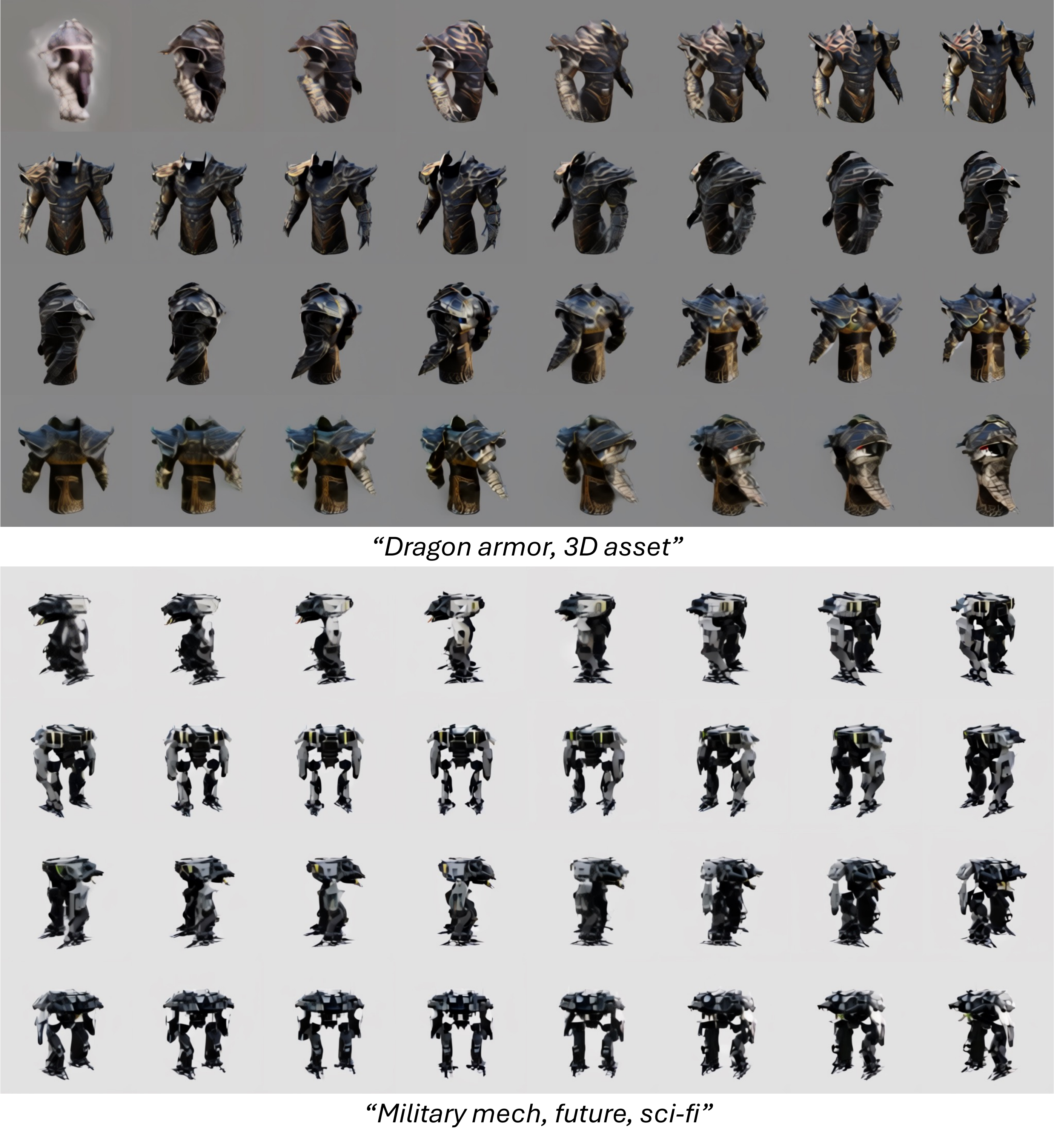}
    \end{center}
      \caption{\textbf{Text-to-32-view results of \methodName}.
    We visualize all 32 frames in this figure for better visibility.
    \methodName shows promising quality and high multi-view consistency and camera coherency, despite generating 32 images in just around 5 seconds.
}
\label{fig:supp_qual_32_a}
\end{figure*}

\section{Additional qualitative results.}

In this section, we provide additional qualitative results of \methodName on full 32 generated views.
The results are shown in~\cref{fig:supp_qual_32_b} to~\cref{fig:supp_qual_32_a}.
\methodName shows promising quality and high multi-view consistency and camera coherency, despite generating 32 images in just around 5 seconds.
As mentioned in~\cref{sec:appendix_drawbacks}, the first image of the generated multi-view images is more prone to blurs, which is strongly visible in the results of the prompt ``Dragon armor, 3D asset".

\clearpage
\clearpage
{
    \small
    \bibliographystyle{ieeenat_fullname}
    \bibliography{main}
}

\end{document}